\documentclass[conference]{IEEEtran}
\IEEEoverridecommandlockouts

\usepackage{cite}
\usepackage{amsmath,amssymb,amsfonts}
\usepackage{algorithmic}
\usepackage{graphicx}
\usepackage{textcomp}
\usepackage{xcolor}
\usepackage{booktabs}
\usepackage{subcaption}
\usepackage{multirow}

\def\BibTeX{{\rm B\kern-.05em{\sc i\kern-.025em b}\kern-.08em
    T\kern-.1667em\lower.7ex\hbox{E}\kern-.125emX}}

\usepackage{url,hyperref,microtype}
\hypersetup{
    colorlinks=true,
    citecolor=blue,
    linkcolor=blue,
    filecolor=magenta,      
    urlcolor=black,
}

\begin{document}

\title{Efficient and Interpretable Body-Based Emotion Recognition with Lightweight Temporal Convolutional Networks}

\author{

\IEEEauthorblockN{Christian Arzate Cruz}
\IEEEauthorblockA{\textit{Honda Research Institute Japan} \\
Wako City, Japan \\
christian.arzate@jp.honda-ri.com}

\and

\IEEEauthorblockN{Stefanos Gkikas}
\IEEEauthorblockA{\textit{Honda Research Institute Japan} \\
Wako City, Japan \\
stefanos.gkikas@jp.honda-ri.com}

\and

\IEEEauthorblockN{Houshyar Asadi}
\IEEEauthorblockA{\textit{Institute for Intelligent Systems Research}\\
\textit{and Innovations (IISRI)} \\
\textit{Deakin University}\\
Geelong, Australia \\
houshyar.asadi@deakin.edu.au}

}

\maketitle

\begin{abstract}
Body-based emotion recognition is important for real-time affective systems, but graph-based skeleton models can be computationally expensive. This paper studies whether lightweight temporal convolutional networks (TCNs) can provide an efficient and interpretable alternative for body-based emotion classification. We evaluate a family of TCN models on DIEM-A and compare them with a graph-based time-series graph (G-TSG) baseline using accuracy, macro-F1, parameter count, and inference latency. Although G-TSG achieves the highest mean performance, TCN-Base remains within $1.58$ accuracy points and $1.25$ macro-F1 points while using $79.18\%$ fewer parameters and reducing classifier latency by approximately $12.5\times$. We also analyze body-region contributions using region-specific TCN models, zero-based occlusion, and G-TSG gradient saliency. The results show that upper-body motion provides the strongest standalone regional cue, that the usefulness of body regions varies across emotions, and that different interpretability methods capture distinct aspects of model behavior. These findings suggest that lightweight TCNs can support efficient body-based emotion recognition while also providing practical insight into how motion cues contribute to classification.
\end{abstract}


\section{Introduction}
Body motion provides important cues for recognizing affective states~\cite{noroozi2018survey}. Human emotions and related affective conditions are expressed not only through facial expressions and speech, but also through posture, gesture, movement dynamics, physiological responses, and whole-body coordination~\cite{iaccino2014left,ruthrof2015body,gkikas_arzate_pain_icmi_2026,gkikas_reface_acii_2026,gkikas_workload_acii_2026}. These cues are especially relevant for interactive systems such as social robots, virtual agents, and affect-aware interfaces, where recognition must often operate under real-time constraints~\cite{arzate_hri_2025, arzate2025when, montiel2026efficient, cruz2024data}.

Recent approaches to body-based emotion recognition have used graph-based neural networks to model the spatial structure of the human body~\cite{shi2020skeleton,ghaleb2021skeleton,zhai2024looking}. These models are well suited for skeleton-based data because they can represent joints as graph nodes and body connections as graph edges. However, graph-convolutional skeleton backbones can introduce nontrivial computational overhead, which can limit their use in real-time interactive settings~\cite{shi2022multiscale,ghaleb2021skeleton}. This motivates the study of lighter architectures that preserve recognition performance while reducing model size and inference latency.

In this paper, we study lightweight temporal convolutional networks (TCNs) for body-based emotion recognition. Our first goal is to evaluate whether small temporal models can approach a stronger graph-based baseline while substantially reducing computational cost. Our second goal is to address the MMAC explainability task by identifying which anatomical regions provide independently useful evidence and which regions a full-body graph model is sensitive to. This distinction matters because an explanation should state both \emph{what} is being measured and \emph{what conclusion it supports}.

To address these goals, we study the DIEM-A challenge setting and compare a graph-based G-TSG baseline with a family of TCN models on the Diverse Intercultural E-Motion Database of Asian Performers (DIEM-A) dataset~\cite{chengasian}. The challenge emphasizes complex social emotions, scenario-dependent expressions, and explainability through prioritized kinematic features or crucial body parts. We therefore evaluate not only accuracy, macro-F1, model size, and latency, but also three complementary explanation questions: which region is sufficient when learned in isolation, what happens when a full-body TCN is perturbed, and which region a full-body G-TSG is locally sensitive to. We further interpret the region results separately for the seven basic and five social emotions used by the challenge.

The contributions of this work are as follows:
\begin{itemize}
    \item We evaluate a family of lightweight TCN models for body-based emotion classification on the DIEM-A dataset and compare them against a graph-based G-TSG baseline.
    \item We show that TCN-Base remains within $1.58$ accuracy points and $1.25$ macro-F1 points of G-TSG while using $79.18\%$ fewer parameters and reducing inference latency by approximately $12.5\times$.
    \item We analyze body-region-specific TCN models to estimate the standalone discriminative contribution of different body regions.
    \item We compare region-specific training, zero-based occlusion, and G-TSG saliency as tests of sufficiency, perturbation sensitivity, and local gradient sensitivity, respectively, and examine how regional evidence varies across basic and social emotions.
\end{itemize}

\section{Related Work}
In this section, we review prior work on body-based emotion recognition, graph-based models for skeleton data, temporal convolutional networks, and interpretability for body-motion analysis. We also clarify how our work differs from prior efficient skeleton models: we focus on a temporal-only classifier for affective body motion and use region-specific training as a practical interpretability tool.

\subsection{Body-Based Emotion Recognition}
Body-based emotion recognition aims to infer affective states from posture, gesture, and movement dynamics. Earlier approaches often relied on hand-crafted geometric and motion features extracted from specific body regions, such as the upper body, hands, head, and torso, as well as full-body motion. For example, affect recognition has been studied using face-body gesture fusion~\cite{gunes2005affect}, body gesture analysis based on head and hand motion~\cite{glowinski2008technique,glowinski2011toward,glowinski2015towards}, and Kinect-based skeletal features, including joint distances, angles, velocities, and accelerations~\cite{saha2014study}. Multimodal settings have also combined body gestures with facial expressions and acoustic cues, showing that body motion provides complementary affective information beyond the face and speech~\cite{kessous2010multimodal}. However, many of these methods rely on shallow geometric or motion-based representations and conventional classifiers, thereby motivating recent work on learned representations for body-based emotion recognition.

The DIEM-A dataset has been used for emotion recognition from body motion~\cite{chengasian}. It provides body-motion data and emotion labels for evaluating models for affective movement analysis. The dataset contains data from $97$ performers, with each performer providing $111$ movement sequences across $12$ emotions, $3$ intensity levels, and $3$ scenarios. This scale makes DIEM-A suitable for studying both recognition performance and how models use body-motion information for classification. In this work, we use DIEM-A to evaluate efficient temporal models and to analyze the standalone contribution of different body regions to body-based emotion recognition.

\subsection{Graph-Based Models for Skeleton Data}
Skeleton-based recognition requires modeling both the body's spatial structure and the temporal evolution of movement. Earlier deep-learning approaches represented skeleton sequences using recurrent models, pseudo-images, or graphs, depending on whether joints were treated as sequences, image-like structures, or graph nodes~\cite{du2015skeleton,yan2018spatial,shi2019skeleton}. Graph-based approaches are especially suitable for skeleton data because they can explicitly encode body joints and their anatomical connections. Spatial-temporal graph convolutional networks have therefore become a common choice for skeleton-based action and emotion recognition~\cite{yan2018spatial,shi2019skeleton,shi2020skeleton}.

However, graph-based models can be computationally demanding. They often combine spatial graph operations with temporal modeling, which can increase model complexity and inference cost. This can be a limitation for real-time applications, especially in interactive systems where affect recognition must be performed with low latency. In this work, we use G-TSG as a graph-based baseline and evaluate whether a simpler TCN architecture can provide a better efficiency--performance trade-off.

\subsection{Temporal Convolutional Networks}
Temporal convolutional networks have been used as efficient alternatives to recurrent models for sequence modeling. In skeleton-based recognition, movement is naturally represented as a temporal sequence of body poses, making temporal modeling essential. Prior work has shown that TCNs can model temporal patterns using causal or dilated convolutions while allowing parallel computation across time steps~\cite{lea2017temporal,bai2018empirical,nan2021comparison}. This property is useful for skeleton-based action and affect recognition because it can reduce the sequential bottleneck of recurrent models while preserving temporal information.

TCNs have also been combined with graph-based spatial modules, where graph operations capture body structure and temporal convolutions model motion dynamics across frames~\cite{yan2018spatial,nan2021comparison}. However, for real-time affective systems, it is also important to evaluate whether temporal models can operate directly on skeleton feature sequences with low computational cost. Recent work has shown that TCN-based models can achieve competitive performance with simpler architectures and faster inference in skeleton-based recognition tasks~\cite{nan2021comparison}. Related compact backbones also include multi-stage temporal models, temporal attention hybrids, and temporal Multilayer Perceptron (MLP)- style sequence models. In this work, we focus on a simple temporal-only TCN because it provides a clear test of how much performance can be retained after removing explicit graph operations. We leave broader lightweight-backbone comparisons to future work.

\subsection{Interpreting Body-Motion Models}
Interpretability in body-based emotion recognition requires more than identifying influential joints. It also requires clarifying what kind of evidence an explanation provides and how that evidence relates to established accounts of expressive movement. The Body Action and Posture coding system characterizes bodily expression across anatomical, form, and functional dimensions~\cite{dael2012body}, while empirical studies show that affective information is conveyed through coordinated patterns of posture and movement distributed across the body~\cite{dael2012emotion}. Cross-cultural research further demonstrates that the perceived meaning of a posture can vary across observer groups~\cite{kleinsmith2006cross}. Consequently, body-region explanations should be interpreted as evidence about a model trained on a particular dataset, population, and elicitation setting; they do not establish fixed correspondences between anatomical regions and emotional states.

Previous work has incorporated anatomical structure directly into model explanations. Ghaleb et al.~\cite{ghaleb2021skeleton}, for example, used body-part attention over the arms, legs, and torso to reveal how a full-body graph model distributes attention across regions. Our analysis complements this approach by examining three distinct properties of regional evidence. Region-specific training estimates how much discriminative information remains available when learning from one region in isolation. Occlusion measures the dependence of a trained full-body model on selected inputs, although zero-masking may introduce implausible skeleton configurations. Gradient saliency captures local sensitivity by quantifying how small input changes affect the class score. Since these methods examine sufficiency, perturbation dependence, and local sensitivity, respectively, agreement among their rankings is neither assumed nor required. Their joint interpretation provides a more precise account of how bodily information supports classification and helps distinguish independently predictive cues from regions that contribute through whole-body coordination.

\section{Method}
In this section, we present the problem formulation and model architectures we use in our studies.

\subsection{Task Formulation}
Given a sequence of body-motion features, the goal is to classify the corresponding emotion label. Let the input sequence be represented as
\begin{equation}
    X = \{x_1, x_2, \ldots, x_T\},
\end{equation}
where $T$ is the number of frames and $x_t$ contains the body-motion features at frame $t$. The model predicts an emotion class
\begin{equation}
    \hat{y} = f(X),
\end{equation}
where $f$ is either the graph-based baseline or a temporal convolutional model.

\subsection{Dataset}
We evaluate models on DIEM-A~\cite{chengasian}, a body-motion dataset for affective movement analysis. DIEM-A contains data from $97$ performers and covers $12$ emotions, $3$ intensity levels, and $3$ scenarios. Following the emotion-recognition setting, each movement sequence is assigned an emotion label and used as one classification sample.

\paragraph{Implementation and Preprocessing}
We follow the standard DIEM-A procedure and official code provided by the dataset authors for preprocessing, fold construction, baseline evaluation, and training configuration. The same preprocessed skeleton sequences and fold definitions are used for G-TSG and all TCN variants, ensuring that comparisons reflect differences in classifiers rather than in preprocessing or data splits. Unless otherwise stated, optimizer settings, training schedule, and evaluation procedures follow the DIEM-A baseline configuration.

We use the provided 6D continuous local joint-rotation representation and convert each sequence into fixed-length clips of $T=64$ frames. The classifier input therefore contains local joint-rotation features, not raw images, 2D keypoints, global translation, velocity, or acceleration features. Joint ordering, sequence length, and root/orientation handling from the official pipeline are kept identical across all folds and models.

\subsection{Models}
We compare a graph-based G-TSG baseline with a family of lightweight temporal convolutional networks (TCNs). All models operate on fixed-length skeleton clips
\begin{equation}
    \mathbf{X} \in \mathbb{R}^{C \times T \times V},
\end{equation}
where $C=6$ is the 6D continuous joint-rotation representation, $T=64$ is the clip length, and $V=25$ is the number of joints in the DIEM-A skeleton. The prediction target is one of $K=12$ emotion classes.

\paragraph{G-TSG baseline}
The G-TSG baseline is implemented as an STGCN++-style spatio-temporal graph convolutional network. It represents the skeleton as a graph in which nodes correspond to body joints and edges represent anatomical connections. The adjacency tensor contains three spatial subsets: self-connections, inward edges, and outward edges. This adjacency tensor is initialized from the skeleton topology and is learnable during training, allowing the model to reweight skeleton connections without using an explicit attention module.

Each G-TSG block applies graph convolution over joints followed by multi-branch temporal convolution over frames. The temporal unit includes branches with kernel size $3$ and dilations $1$, $2$, $3$, and $4$, together with max-pooling and $1 \times 1$ convolution branches. This design allows the model to capture short- and long-range motion patterns. The full network stacks ten spatio-temporal blocks, with four blocks at $64$ channels, three blocks at $128$ channels, and three blocks at $256$ channels. Temporal downsampling is applied at the transitions from $64$ to $128$ channels and from $128$ to $256$ channels. Residual connections are used throughout the network, except when the input and output dimensions differ. After the final block, global average pooling over time and joints is followed by dropout and a linear classifier.

\paragraph{Lightweight TCN models}
The TCN models remove explicit graph convolution and treat each skeleton clip as a multivariate temporal signal. For each frame, the joint dimension is flattened, reshaping the input from $\mathbf{X} \in \mathbb{R}^{C \times T \times V}$ to $\mathbf{X}_{\mathrm{TCN}} \in \mathbb{R}^{(CV) \times T}$. The model then applies input batch normalization, a $1 \times 1$ projection to a hidden width $d$, and a stack of residual dilated 1D temporal convolution blocks. Each block uses kernel size $3$, batch normalization, ReLU, dropout, and a residual connection. After the final block, temporal average pooling produces a clip-level embedding, which is passed through dropout and a linear classifier.

The five TCN variants differ only in their hidden widths and dilation schedules, as shown in Table~\ref{tab:tcn_architectures}. All variants use dropout $0.5$.

\begin{table}[htb]
\centering
\caption{Architecture settings for the lightweight TCN family.}
\label{tab:tcn_architectures}
\begin{tabular}{lcc}
\toprule
Model & Hidden width $d$ & Dilations \\
\midrule
TCN-S & 64  & $[1,2,4]$ \\
TCN-M & 96  & $[1,2,4,8]$ \\
TCN-Base & 128 & $[1,2,4,8]$ \\
TCN-L & 192 & $[1,2,4,8]$ \\
TCN-XL & 128 & $[1,2,4,8,16]$ \\
\bottomrule
\end{tabular}
\end{table}

\paragraph{Body-region TCN models}
For the body-region analysis, we train separate TCN-Base models using restricted body-region inputs. These models use the same architecture and training protocol as the full-body TCN-Base model, but the input is limited to a particular anatomical region. This evaluates the classification utility of each region when used alone. In contrast, the G-TSG saliency analysis measures inference-time sensitivity of a full-body graph model. Therefore, the region-specific TCN models and the G-TSG saliency results provide complementary views of how body-motion information supports emotion recognition.

\paragraph{Body-region definitions}
We define body regions from the $25$-joint DIEM-A skeleton using zero-based Python joint indices. The four non-overlapping regions are head, torso, arms/hands, and legs. Head contains joints $[6,7,8]$ (Neck, Neck1, Head). Torso contains joints $[0,1,2,3,4,5]$ (Root, Hips, Spine, Spine1, Spine2, Spine3). Arms/hands contain joints $[9,10,11,12,13,14,15,16]$ (RightShoulder, RightArm, RightForeArm, RightHand, LeftShoulder, LeftArm, LeftForeArm, LeftHand). Legs contain joints $[17,18,19,20,21,22,23,24]$ (RightUpLeg, RightLeg, RightFoot, RightToeBase, LeftUpLeg, LeftLeg, LeftFoot, LeftToeBase). The upper body is the union of the head, torso, and arms/hands, corresponding to joints $0$--$16$. For region-specific TCN training and G-TSG saliency, the lower body includes the Root and Hips, plus the leg joints, corresponding to $[0,1,17,18,19,20,21,22,23,24]$. The same mapping is used for region-specific training and saliency aggregation; for occlusion, we report the four direct non-overlapping region masks.

\subsection{Evaluation Protocol}
We evaluate models using leave-performer-out cross-validation, which is the standard protocol for DIEM-A. Accuracy and macro-F1 are reported as mean $\pm$ standard deviation across $10$ folds. Each fold corresponds to a different group of performers, so test performers are not seen during training. Because the task has $12$ emotion classes, chance accuracy is approximately $8.33\%$. We report trainable parameters, multiply--accumulate operations (MACs), estimated memory footprint, and inference latency to evaluate computational efficiency.

Inference latency is measured with batch size $1$ using a single fixed-shape DIEM-A clip with $C=6$, $T=64$, and $V=25$. All models are evaluated in FP32 mode after $50$ warm-up forward passes, followed by $200$ timed forward passes. For CUDA timing, we synchronize the device before starting each timed iteration and immediately after each forward pass using \texttt{torch.cuda.synchronize()}, ensuring that asynchronous GPU execution does not bias the measurements. The benchmark metadata are: RTX 3090 24GB GPU, Intel i9-10900F CPU, 62.71 GiB RAM, Linux 5.15, Python 3.11.15, PyTorch 2.11.0+cu130, CUDA 13.0, cuDNN 91900, FP32, \texttt{cudnn.benchmark=False}, and \texttt{cudnn.deterministic=False}. The reported latency reflects only the classifier stage; upstream pose estimation or motion-capture processing is not included.

\subsection{Body-Region Analysis}
\label{sec:body_region_analysis}
To analyze how body-motion information supports model decisions, we use three complementary analyses.
First, we train body-region-specific TCN-Base models. Each model is trained and tested using only one input region: full body, upper body, lower body, arms/hands, legs, or torso. This analysis estimates the region-only cue of each body region under matched training and testing conditions.

Second, we perform zero-based occlusion using the trained full-body TCN-Base model. In this analysis, selected body regions are replaced with zeros during inference. This analysis is treated as diagnostic because zero-masking introduces a train-test distribution shift.

Third, we compare the region-specific TCN results with G-TSG input-gradient saliency. This comparison allows us to distinguish between standalone regional discriminative power and model sensitivity during full-body graph-based inference. For saliency, we compute the absolute gradient of the ground-truth class logit with respect to the input sequence, aggregate the magnitudes across channels and time, average over joints within each body region to reduce region-size bias, and normalize the resulting region scores so that they sum to $100\%$ per sample. We then average the normalized scores across samples and report aggregate and per-emotion saliency values.

\section{Results}
We first compare G-TSG and the TCN family in terms of performance and efficiency, and then analyze body-region evidence using region-specific TCNs, zero-based occlusion, and G-TSG saliency.

\subsection{Performance and Model Size Comparison}
Table~\ref{tab:diema_tcn_family_latency} compares the graph-based G-TSG baseline with the TCN family. G-TSG achieves the highest overall accuracy and macro-F1. However, TCN-Base reaches a similar performance range while using substantially fewer parameters and lower latency. Compared with G-TSG, TCN-Base reduces the number of parameters by $79.18\%$ and decreases inference latency from $13.85$ ms to $1.11$ ms per clip. This corresponds to an approximately $12\times$ latency reduction, while remaining within $1.58$ accuracy points and $1.25$ macro-F1 points of the baseline.

These results suggest that lightweight temporal convolutional models provide a favorable trade-off between recognition performance and computational efficiency. TCN-Base uses $79.18\%$ fewer parameters, $98.03\%$ fewer MACs, and an estimated $96.36\%$ less memory than G-TSG, while reducing latency by $12.51\times$. Using paired fold-level comparisons across the $10$ leave-performer-out folds, G-TSG achieves higher mean accuracy than TCN-Base by $1.58$ percentage points, but this difference is not statistically significant (Wilcoxon signed-rank test: $W=17.0$, $p=0.322$; paired $t$-test: $t(9)=1.196$, $p=0.262$). Similarly, G-TSG exceeds TCN-Base in macro-F1 by $1.25$ points, but the paired difference is not statistically significant (Wilcoxon signed-rank test: $W=22.0$, $p=0.625$; paired $t$-test: $t(9)=0.809$, $p=0.440$). Under these paired fold-level tests, we did not detect a statistically significant performance difference between G-TSG and TCN-Base. Together with its substantially lower computational cost, these results support considering TCN-Base a practical low-latency alternative, while not establishing statistical equivalence or non-inferiority.

\begin{table*}[t]
\centering
\footnotesize
\setlength{\tabcolsep}{3.5pt}
\renewcommand{\arraystretch}{0.95}
\caption{DIEM-A performance and computational efficiency for G-TSG and the lightweight TCN family. Accuracy and macro-F1 are mean $\pm$ standard deviation across folds. Latency is measured per clip, with a batch size of $1$ on CUDA. MACs count convolutional and linear layers; for G-TSG, dense skeleton-adjacency multiplications inside graph convolution are also included. Italicized memory values are estimated model-tensor and activation footprints because CUDA allocation profiling was unavailable in the current run.}
\label{tab:diema_tcn_family_latency}
\begin{tabular}{lrrrrrr}
\toprule
Model & Params & MACs/clip & Mem. (MB) & Test Acc. (\%) & Test F1 (\%) & Latency (ms) \\
\midrule
G-TSG & 1,376,751 & 914.92M & \textit{37.91} & \textbf{27.11 $\pm$ 3.67} & \textbf{25.21 $\pm$ 4.49} & 13.851 $\pm$ 0.177 \\
\midrule
TCN-S & 61,176 & 3.76M & \textit{0.35} & 24.45 $\pm$ 2.93 & 22.23 $\pm$ 3.09 & \textbf{0.878 $\pm$ 0.003} \\
TCN-M & 165,912 & 10.36M & \textit{0.85} & 25.11 $\pm$ 2.48 & 23.68 $\pm$ 2.44 & \underline{1.096 $\pm$ 0.011} \\
TCN-Base & 286,648 & 18.01M & \textit{1.38} & \underline{25.53 $\pm$ 3.10} & \underline{23.96 $\pm$ 2.88} & 1.107 $\pm$ 0.020 \\
TCN-L & 626,424 & 39.59M & \textit{2.83} & 23.81 $\pm$ 2.30 & 22.50 $\pm$ 2.21 & 1.120 $\pm$ 0.020 \\
TCN-XL & 352,952 & 22.20M & \textit{1.70} & 25.03 $\pm$ 2.53 & 23.09 $\pm$ 2.50 & 1.326 $\pm$ 0.014 \\
\bottomrule
\end{tabular}
\end{table*}

We use TCN-Base for the body-region analysis because it provides the best overall performance--efficiency trade-off among the TCN variants. The regional-training and occlusion analyses were conducted in separate controlled experimental runs; consequently, their full-body reference values differ slightly from the primary TCN-Base results reported in Table~\ref{tab:diema_tcn_family_latency}. All performance drops and statistical comparisons are therefore computed relative to the corresponding full-body reference from the same run.

\subsection{Body-Region-Specific TCN Analysis}
After establishing TCN-Base as the most balanced TCN variant, we use it to analyze body-region contribution. Table~\ref{tab:tcn_region_performance} shows that the upper-body model is closest to the full-body model, with drops of only $1.76$ accuracy points and $2.03$ macro-F1 points. The upper-body macro-F1 difference is not statistically significant after Holm correction, whereas the macro-F1 differences for arms/hands, lower body, legs, and torso are statistically significant. 

These results suggest that upper-body motion contains the strongest standalone regional cue for body-based emotion recognition in this setting. However, no single region outperforms the full-body model, indicating that emotion recognition still benefits from distributed whole-body motion patterns.

\begin{table}[t]
\centering
\caption{Body-region TCN-Base results on DIEM-A. Drops are relative to the full-body model; $p$-values are Holm-adjusted paired Wilcoxon tests on fold-level macro-F1.}
\label{tab:tcn_region_performance}
\begin{tabular}{lrrrrr}
\toprule
Input Region & Acc. & Acc Drop & F1 & F1 Drop & $p_{\mathrm{Holm}}$ (F1) \\
\midrule
Full body & \textbf{24.90} & 0.00 & \textbf{23.69} & 0.00 & -- \\
Upper body & \underline{23.14} & \textbf{1.76} & \underline{21.66} & \textbf{2.03} & 0.0645 \\
Arms/hands & 20.59 & 4.31 & 19.09 & 4.59 & 0.0098 \\
Lower body & 19.38 & 5.52 & 17.51 & 6.17 & 0.0098 \\
Legs & 19.18 & 5.72 & 17.39 & 6.30 & 0.0098 \\
Torso & 17.43 & 7.47 & 15.27 & 8.42 & 0.0098 \\
\bottomrule
\end{tabular}
\end{table}

\subsection{Per-Emotion Analysis of Body-Region Contributions} 
\begin{figure*}[t]
\centering
\includegraphics[width=\textwidth]{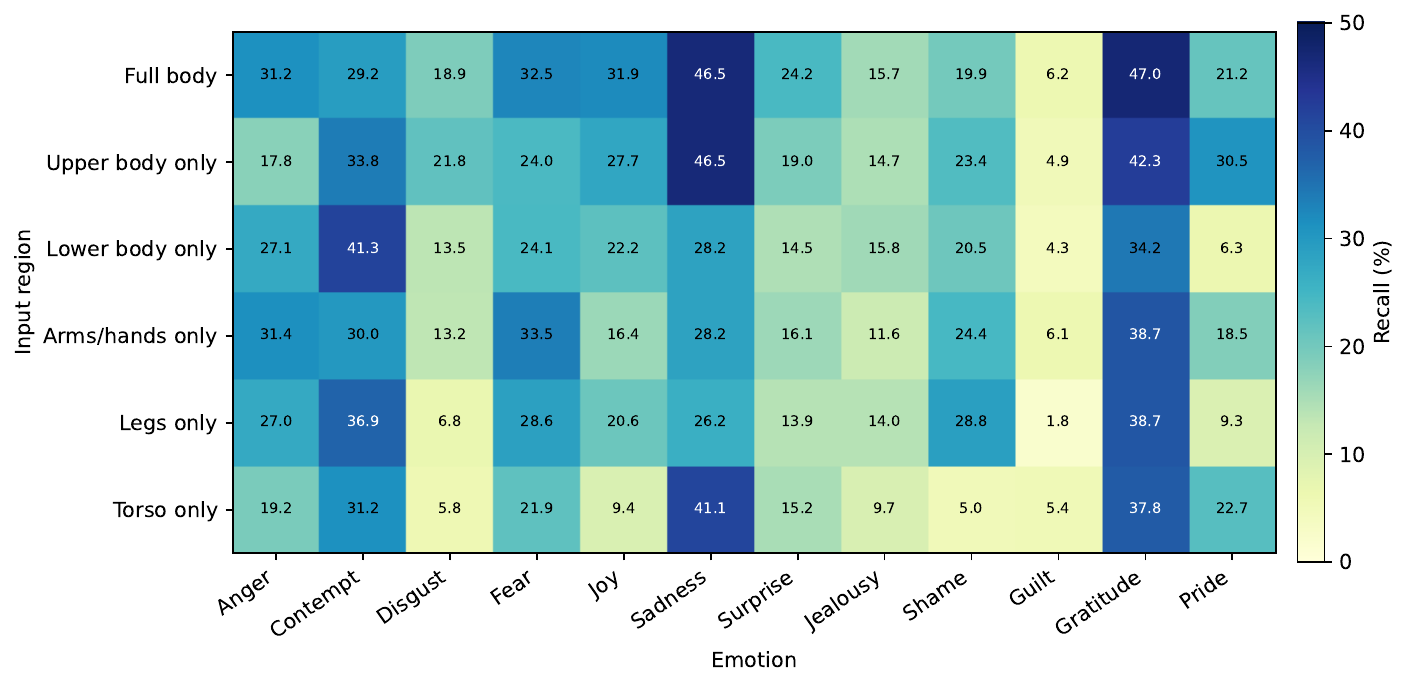}
\caption{Per-emotion recall for TCN-Base models trained with different body-region inputs. Emotion-relevant information is not uniformly distributed across the body: upper-body motion is strongest overall, while the arms/hands, legs, and torso retain comparatively useful evidence for particular emotion classes.}
\label{fig:region_emotion_heatmap}
\end{figure*}

The global region results show that upper-body motion is the strongest region-only cue overall, but this does not mean that every emotion relies on the same body region. Because the 12-way task has modest absolute performance, we report per-emotion recall in Fig.~\ref{fig:region_emotion_heatmap} to examine how predictive information varies across body regions and emotion classes; a full confusion-matrix analysis is left to supplementary material. 

The heatmap indicates that emotion-relevant information is distributed differently across body regions. Upper-body motion performs well for several emotions and achieves the highest macro recall among the region-only models. Arms/hands achieve comparatively high recall for anger and fear. Lower-body and leg inputs retain comparatively useful information for contempt and shame. Torso-only input is weaker overall, but it still provides useful information for some emotions, such as sadness and gratitude.

These results show that body-region contribution is emotion-dependent and that aggregate accuracy and macro-F1 can conceal substantial differences across emotion classes.

Additionally, we evaluated zero-based occlusion using the trained full-body TCN-Base model. In this analysis, selected body regions were replaced with zeros during inference. Table~\ref{tab:tcn_global_occlusion} shows that replacing any body region with zeros causes a large performance drop. However, the drops are similar across conditions, making it difficult to identify a clear hierarchy of body-region importance. 

Zero-based occlusion likely introduces a strong train--test distribution shift because the model is trained on complete skeletons but tested with invalid zeroed body configurations. Therefore, we treat this result as diagnostic: it shows that the full-body model depends on complete body structure, but it does not provide a reliable ranking of body-region importance.

\begin{table}[t]
\centering
\caption{Zero-based occlusion diagnostic for TCN-Base on DIEM-A. Drops are relative to full-body inference.}
\label{tab:tcn_global_occlusion}
\begin{tabular}{lrrrr}
\toprule
Condition & Acc. & Acc Drop & F1 & F1 Drop \\
\midrule
No occlusion & 27.08 & 0.00 & 24.61 & 0.00 \\
Occlude head & 8.22 & 18.87 & 1.27 & 23.34 \\
Occlude arms/hands & 9.84 & 17.25 & 3.33 & 21.28 \\
Occlude torso & 8.33 & 18.75 & 1.28 & 23.32 \\
Occlude legs & 9.38 & 17.71 & 2.78 & 21.82 \\
\bottomrule
\end{tabular}
\end{table}

\subsection{Comparing Body-Region Performance and Model Saliency}
Finally, Table~\ref{tab:tcn_gtsg_region_comparison} compares body-region-specific TCN performance with G-TSG input-gradient saliency. The saliency scores are normalized by region size before averaging, as described in Section~\ref{sec:body_region_analysis}. These two analyses should not be interpreted as measuring the same quantity. Region-specific TCN performance measures the discriminative contribution of a body region in isolation, with the model trained and tested on only that region. In contrast, G-TSG saliency measures the graph model's sensitivity to each region during full-body inference.

The comparison shows that the torso has the lowest standalone TCN performance but the highest G-TSG saliency. This does not necessarily indicate a contradiction. Instead, it suggests that the torso may be weak as an isolated cue but important as a structural or contextual anchor for the graph-based model. Similarly, arms/hands provide stronger standalone TCN performance than torso, but lower G-TSG saliency. These differences show why interpretability results should be interpreted according to the question each method answers.

\begin{table}[t]
\centering
\caption{Comparison of body-region TCN performance and G-TSG saliency.}
\label{tab:tcn_gtsg_region_comparison}
\begin{tabular}{lrrrr}
\toprule
Body Region & TCN Acc. & TCN F1 & F1 Drop & G-TSG Saliency \\
\midrule
Torso & 17.4 & 15.3 & 8.4 & 47.8 \\
Arms/hands & 20.6 & 19.1 & 4.6 & 13.2 \\
Legs & 19.2 & 17.4 & 6.3 & 22.2 \\
\bottomrule
\end{tabular}
\end{table}

\begin{figure*}[t]
\centering
\includegraphics[width=\textwidth]{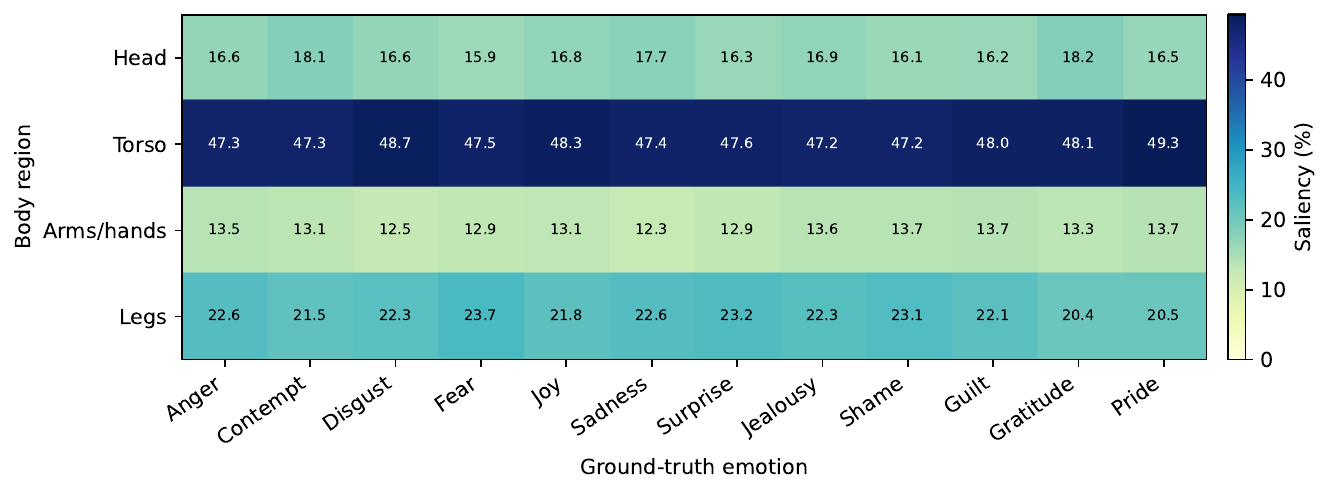}
\caption{G-TSG input-gradient saliency by body region and ground-truth emotion. Saliency is consistently highest for the torso across emotions, followed by the legs, head, and arms/hands.}
\label{fig:nonoverlap_heatmap}
\end{figure*}

In addition to the aggregate saliency values, we also visualize G-TSG input-gradient saliency by emotion in Figure~\ref{fig:nonoverlap_heatmap}. The heatmap shows a consistent pattern across emotions: G-TSG assigns the highest saliency to the torso, followed by the legs, head, and arms/hands. This indicates that the G-TSG output is most locally sensitive to torso-related inputs under the adopted gradient-saliency measure.

\section{Discussion}
We discuss computational efficiency and interpretability in body-based emotion recognition. Region-specific training, occlusion, and gradient saliency provide complementary views of regional discriminative value, model dependence, and prediction sensitivity.

\subsection{Efficiency for Real-Time Affective Systems}
The paired fold-level tests found no statistically significant difference between G-TSG and TCN-Base in accuracy or macro-F1. Although this does not establish equivalence, TCN-Base offers a substantially more efficient low-latency alternative.

The comparison is limited to one graph baseline and temporal-only TCNs. It therefore supports TCN-Base as an efficient alternative to G-TSG, not as the optimal lightweight skeleton architecture. Parameter-matched GCNs, compact Transformers, temporal MLPs, and shallow kinematic baselines remain important comparisons.

\subsection{Emotion-Specific Body-Region Evidence}
The upper body provides the strongest standalone regional evidence and is the only region-specific model whose macro-F1 difference from the full-body model is not statistically significant after Holm correction. This result is consistent with perceptual and computational studies emphasizing upper-body and arm cues~\cite{kleinsmith2012affective,ghaleb2021skeleton}, although the per-emotion analysis reveals a more differentiated pattern. Arms/hands achieve comparatively high recall for anger and fear, legs for shame, and the torso for sadness and gratitude. These findings are also compatible with body-coding research showing that emotion-related information is distributed across anatomical actions and movement form~\cite{dael2012body,dael2012emotion}.

The distinction between basic and social emotions provides an additional perspective on the regional results. The full-body model performs better on basic emotions, whereas the upper-body model preserves a larger proportion of the recall for social emotions. This does not imply that social emotions are inherently expressed through the upper body. Within DIEM-A, upper-body dynamics retain comparatively useful evidence for classes such as gratitude and pride, while guilt remains difficult across all regional models. These results illustrate how region-level analyses can clarify where a classifier obtains evidence for challenging emotion categories without turning model behavior into a general psychological claim.

\subsection{What Each Explanation Supports}
The three analyses form an explicit claim hierarchy. Region-only training tests \emph{sufficiency}: how well a new classifier can solve the task from one anatomical subset. Occlusion tests \emph{perturbation dependence}, but zeroing joints creates invalid poses and therefore provides only a diagnostic stress test. Input gradients test \emph{local sensitivity} of G-TSG around each observed sequence; they do not measure standalone predictive value. This separation explains why the methods need not agree.

The torso is the clearest example. It is weak in isolation yet receives the largest G-TSG gradient share. The graph model may use the torso as a reference frame for coordinated limb motion, while the lower performance of the torso-only TCN suggests that torso motion becomes more informative when interpreted together with complementary limb trajectories. Conversely, arms/hands are more sufficient in isolation but receive lower graph-model saliency. Compared with body-part attention in Ghaleb et al.~\cite{ghaleb2021skeleton}, our results show why an anatomical heatmap should be paired with an intervention or restricted-input test before it is interpreted as feature importance. The contribution is therefore a triangulated explanation: agreement strengthens a regional claim, while disagreement reveals architecture-dependent use of the skeleton.

\section{Conclusion}
We evaluated lightweight temporal convolutional networks for body-based emotion recognition on DIEM-A. Although G-TSG achieved the highest performance, TCN-Base remained competitive while using $79.18\%$ fewer parameters and approximately $12.5\times$ lower latency, supporting its use in compact, real-time affective systems.

We also compared regional sufficiency, occlusion sensitivity, and gradient saliency. Upper-body motion provided the strongest standalone evidence, whereas the torso was weak in isolation but most salient to G-TSG, showing that explanations depend on both the method and architecture. These findings demonstrate the value of combining complementary explanation methods and motivate future culture- and scenario-specific analyses.

\section*{Ethical Impact Statement}
Body-based emotion recognition can support affect-aware robots, virtual agents, and interactive interfaces, but it also creates risks if emotion predictions are treated as objective readings of a person's internal state. Body motion is culturally shaped, context-dependent, and affected by individual differences, so misclassification may lead to incorrect assumptions about a person's feelings, intentions, or needs. For this reason, body-only emotion recognition should be used only as a supportive signal, communicated with uncertainty, and avoided in high-stakes settings such as hiring, education assessment, policing, healthcare diagnosis, or access control.

Ethical deployment also requires attention to privacy, consent, fairness, and responsible data use. The DIEM-A dataset was obtained through official access from the dataset providers and used only for research purposes. Even when skeleton features are less identifiable than raw video, they can still reveal sensitive behavioral patterns and should be collected, stored, and processed with appropriate safeguards. Because this study uses controlled DIEM-A data from Asian performers, the results should not be assumed to generalize across cultures, ages, body types, abilities, or real-world interaction contexts. Future systems should be evaluated for cross-cultural robustness, demographic bias, transparency, user consent, and privacy-preserving operation before being deployed in real interactive environments.

\bibliographystyle{IEEEtran}
\bibliography{bibliography}

@inproceedings{chengasian,
  author    = {Cheng, Miao and Tseng, Chia-huei and Fujiwara, Ken and Schneider, Victor and Kitamura, Yoshifumi},
  title     = {Asian Emotional Body Movement Database: Diverse Intercultural E-Motion Database of Asian Performers ({DIEM-A})},
  booktitle = {2025 13th International Conference on Affective Computing and Intelligent Interaction ({ACII})},
  year      = {2025},
  publisher = {IEEE}
}

@article{noroozi2018survey,
  title={Survey on emotional body gesture recognition},
  author={Noroozi, Fatemeh and Corneanu, Ciprian Adrian and Kami{\'n}ska, Dorota and Sapi{\'n}ski, Tomasz and Escalera, Sergio and Anbarjafari, Gholamreza},
  journal={IEEE transactions on affective computing},
  volume={12},
  number={2},
  pages={505--523},
  year={2018},
  publisher={IEEE}
}

@book{iaccino2014left,
  title={Left brain-right brain differences: Inquiries, evidence, and new approaches},
  author={Iaccino, James F},
  year={2014},
  publisher={Psychology Press}
}

@article{ruthrof2015body,
  title={The body in language},
  author={Ruthrof, Horst},
  year={2015},
  publisher={Bloomsbury Publishing}
}

@INPROCEEDINGS{arzate_hri_2025,
  author={Arzate Cruz, Christian and Montiel-Vázquez, Edwin C. and Maeda, Chikara and Lam, Darryl and Gomez, Randy},
  booktitle={2025 20th ACM/IEEE International Conference on Human-Robot Interaction (HRI)}, 
  title={Empathetic Robots Using Empathy Classifiers in HRI Settings}, 
  year={2025},
  volume={},
  number={},
  pages={1211-1215},
  doi={10.1109/HRI61500.2025.10974139}
  }

@article{montiel2026efficient,
  title={Efficient Emotion-Aware Iconic Gesture Prediction for Robot Co-Speech},
  author={Montiel-Vazquez, Edwin C and Cruz, Christian Arzate and Gkikas, Stefanos and Kassiotis, Thomas and Giannakakis, Giorgos and Gomez, Randy},
  journal={arXiv preprint arXiv:2604.11417},
  year={2026}
}

@article{shi2020skeleton,
  title={Skeleton-based emotion recognition based on two-stream self-attention enhanced spatial-temporal graph convolutional network},
  author={Shi, Jiaqi and Liu, Chaoran and Ishi, Carlos Toshinori and Ishiguro, Hiroshi},
  journal={Sensors},
  volume={21},
  number={1},
  pages={205},
  year={2020},
  publisher={MDPI}
}

@inproceedings{ghaleb2021skeleton,
  title={Skeleton-based explainable bodily expressed emotion recognition through graph convolutional networks},
  author={Ghaleb, Esam and Mertens, Andr{\'e} and Asteriadis, Stylianos and Weiss, Gerhard},
  booktitle={2021 16th IEEE International Conference on Automatic Face and Gesture Recognition (FG 2021)},
  pages={1--8},
  year={2021},
  organization={IEEE}
}

@article{zhai2024looking,
  title={Looking into gait for perceiving emotions via bilateral posture and movement graph convolutional networks},
  author={Zhai, Yingjie and Jia, Guoli and Lai, Yu-Kun and Zhang, Jing and Yang, Jufeng and Tao, Dacheng},
  journal={IEEE Transactions on Affective Computing},
  volume={15},
  number={3},
  pages={1634--1648},
  year={2024},
  publisher={IEEE}
}

@article{shi2022multiscale,
  title={Multiscale 3D-shift graph convolution network for emotion recognition from human actions},
  author={Shi, Henglin and Peng, Wei and Chen, Haoyu and Liu, Xin and Zhao, Guoying},
  journal={IEEE Intelligent Systems},
  volume={37},
  number={4},
  pages={103--110},
  year={2022},
  publisher={IEEE}
}

@inproceedings{gunes2005affect,
  title={Affect recognition from face and body: early fusion vs. late fusion},
  author={Gunes, Hatice and Piccardi, Massimo},
  booktitle={2005 IEEE international conference on systems, man and cybernetics},
  volume={4},
  pages={3437--3443},
  year={2005},
  organization={IEEE}
}

@inproceedings{glowinski2008technique,
  title={Technique for automatic emotion recognition by body gesture analysis},
  author={Glowinski, Donald and Camurri, Antonio and Volpe, Gualtiero and Dael, Nele and Scherer, Klaus},
  booktitle={2008 IEEE Computer society conference on computer vision and pattern recognition workshops},
  pages={1--6},
  year={2008},
  organization={IEEE}
}

@inproceedings{glowinski2015towards,
  author    = {D. Glowinski and M. Mortillaro and K. Scherer and N. Dael and G. Volpe and A. Camurri},
  title     = {Towards a minimal representation of affective gestures},
  booktitle = {2015 International Conference on Affective Computing and Intelligent Interaction (ACII)},
  pages     = {498--504},
  year      = {2015},
  publisher = {IEEE}
}

@article{glowinski2011toward,
  title={Toward a minimal representation of affective gestures},
  author={Glowinski, Donald and Dael, Nele and Camurri, Antonio and Volpe, Gualtiero and Mortillaro, Marcello and Scherer, Klaus},
  journal={IEEE Transactions on Affective Computing},
  volume={2},
  number={2},
  pages={106--118},
  year={2011},
  publisher={IEEE}
}

@inproceedings{saha2014study,
  title={A study on emotion recognition from body gestures using Kinect sensor},
  author={Saha, Sriparna and Datta, Shreyasi and Konar, Amit and Janarthanan, Ramadoss},
  booktitle={2014 international conference on communication and signal processing},
  pages={056--060},
  year={2014},
  organization={IEEE}
}

@article{kessous2010multimodal,
  title={Multimodal emotion recognition in speech-based interaction using facial expression, body gesture and acoustic analysis},
  author={Kessous, Loic and Castellano, Ginevra and Caridakis, George},
  journal={Journal on Multimodal User Interfaces},
  volume={3},
  number={1},
  pages={33--48},
  year={2010},
  publisher={Springer}
}

@inproceedings{du2015skeleton,
  title={Skeleton based action recognition with convolutional neural network},
  author={Du, Yong and Fu, Yun and Wang, Liang},
  booktitle={2015 3rd IAPR Asian conference on pattern recognition (ACPR)},
  pages={579--583},
  year={2015},
  organization={IEEE}
}

@inproceedings{yan2018spatial,
  title={Spatial temporal graph convolutional networks for skeleton-based action recognition},
  author={Yan, Sijie and Xiong, Yuanjun and Lin, Dahua},
  booktitle={Proceedings of the AAAI conference on artificial intelligence},
  volume={32},
  number={1},
  year={2018}
}

@inproceedings{shi2019skeleton,
  title={Skeleton-based action recognition with directed graph neural networks},
  author={Shi, Lei and Zhang, Yifan and Cheng, Jian and Lu, Hanqing},
  booktitle={Proceedings of the IEEE/CVF conference on computer vision and pattern recognition},
  pages={7912--7921},
  year={2019}
}

@inproceedings{gkikas_arzate_pain_icmi_2026,
  title={{A Unified Tokenization Framework for Pain Recognition using Heterogeneous 3D Modalities}},
  author={Stefanos Gkikas and Christian Arzate Cruz and Valentina Becchetti and Muhammad Umar Khan and Alessandro Giuseppi and Raul Fernandez Rojas},
  booktitle={Proceedings of the 28th ACM International Conference on Multimodal Interaction},
  year={2026},
  location={Napoli, Italy},
  publisher={Association for Computing Machinery}
}

@inproceedings{gkikas_reface_acii_2026,
  title={{ReFace: Reorganizing Facial Spatiotemporal Representations for Improved Pain Assessment}},
  author={Stefanos Gkikas and Yu Fang and Christian Arzate Cruz and Muhammad Umar Khan and Raul Fernandez Rojas},
  booktitle={2026 14th International Conference on Affective Computing and Intelligent Interaction (ACII)},
  year={2026},
  location={Puebla, Mexico},
  publisher={IEEE}
}

@inproceedings{gkikas_workload_acii_2026,
  title={{Towards a Unified Modality-Agnostic Multimodal Framework for Cognitive Workload Assessment}},
  author={Stefanos Gkikas and Christian Arzate Cruz and Calvin Joseph and Giorgos Giannakakis and Raul Fernandez Rojas},
  booktitle={2026 14th International Conference on Affective Computing and Intelligent Interaction (ACII)},
  year={2026},
  location={Puebla, Mexico},
  publisher={IEEE}
}

@inproceedings{lea2017temporal,
  title={Temporal convolutional networks for action segmentation and detection},
  author={Lea, Colin and Flynn, Michael D and Vidal, Rene and Reiter, Austin and Hager, Gregory D},
  booktitle={proceedings of the IEEE Conference on Computer Vision and Pattern Recognition},
  pages={156--165},
  year={2017}
}

@article{bai2018empirical,
  title={An empirical evaluation of generic convolutional and recurrent networks for sequence modeling},
  author={Bai, Shaojie and Kolter, J Zico and Koltun, Vladlen},
  journal={arXiv preprint arXiv:1803.01271},
  year={2018}
}

@article{nan2021comparison,
  title={Comparison between recurrent networks and temporal convolutional networks approaches for skeleton-based action recognition},
  author={Nan, Mihai and Tr{\u{a}}sc{\u{a}}u, Mihai and Florea, Adina Magda and Iacob, Cezar C{\u{a}}t{\u{a}}lin},
  journal={Sensors},
  volume={21},
  number={6},
  pages={2051},
  year={2021},
  publisher={MDPI}
}

@article{kleinsmith2006cross,
  title={Cross-cultural differences in recognizing affect from body posture},
  author={Kleinsmith, Andrea and De Silva, P Ravindra and Bianchi-Berthouze, Nadia},
  journal={Interacting with computers},
  volume={18},
  number={6},
  pages={1371--1389},
  year={2006},
  publisher={Oxford University Press Oxford, UK}
}

@article{kleinsmith2012affective,
  title={Affective body expression perception and recognition: A survey},
  author={Kleinsmith, Andrea and Bianchi-Berthouze, Nadia},
  journal={IEEE Transactions on Affective Computing},
  volume={4},
  number={1},
  pages={15--33},
  year={2012},
  publisher={IEEE}
}

@article{dael2012body,
  title={The body action and posture coding system (BAP): Development and reliability},
  author={Dael, Nele and Mortillaro, Marcello and Scherer, Klaus R},
  journal={Journal of Nonverbal Behavior},
  volume={36},
  number={2},
  pages={97--121},
  year={2012},
  publisher={Springer}
}

@article{dael2012emotion,
  title={Emotion expression in body action and posture.},
  author={Dael, Nele and Mortillaro, Marcello and Scherer, Klaus R},
  journal={Emotion},
  volume={12},
  number={5},
  pages={1085},
  year={2012},
  publisher={American Psychological Association}
}

@inproceedings{arzate2025when,
  title={When and How to Express Empathy in Human-Robot Interaction Scenarios},
  author={Arzate Cruz, Christian and Montiel-Vazquez, Edwin C and Maeda, Chikara and Gomez, Randy},
  booktitle={2025 34th IEEE International Conference on Robot and Human Interactive Communication (RO-MAN)},
  pages={1070--1077},
  year={2025},
  organization={IEEE}
}

@inproceedings{cruz2024data,
  title={Data augmentation for 3dmm-based arousal-valence prediction for hri},
  author={Arzate Cruz, Christian and Sechayk, Yotam and Igarashi, Takeo and Gomez, Randy},
  booktitle={2024 33rd IEEE International Conference on Robot and Human Interactive Communication (ROMAN)},
  pages={2015--2022},
  year={2024},
  organization={IEEE}
}

\end{document}